# Data-Driven Extreme Response Estimation

Samuel J. Edwards and Michael D. Levine

Naval Surface Warfare Center, Carderock Division

**ABSTRACT**
A method to rapidly estimate extreme ship response events is developed in this paper. The method involves training by a Long Short-Term Memory (LSTM) neural network to correct a lower-fidelity hydrodynamic model to the level of a higher-fidelity simulation. More focus is placed on larger responses by isolating the time-series near peak events identified in the lower-fidelity simulations and training on only the shorter time-series around the large event. The method is tested on the estimation of pitch time-series maxima in Sea State 5 (significant wave height of 4.0 meters and modal period of 15.0 seconds,) generated by a lower-fidelity hydrodynamic solver known as SimpleCode and a higher-fidelity tool known as the Large Amplitude Motion Program (LAMP). The results are also compared with an LSTM trained without special considerations for large events.

1.  **INTRODUCTION AND BACKGROUND**

The ship design process involves many challenges but one of the most important is the consideration of extreme events. Knowledge of ocean wave conditions that may cause extreme responses is imperative for safe operation of a ship though the impact on day-to-day routine operations in calm seas is small. Obtaining this knowledge involves understanding the stochastic nature of the seaway conditions, non-linear hydrodynamics in waves, and the corresponding non-linear vessel dynamics. Consequently, extreme responses and conditions are difficult to predict due to the stochastic nature and nonlinearity of the events.

The most straightforward approach to estimating extremes of stochastic non-linear systems is through Monte Carlo simulations. However, for most tools of reasonable fidelity, the computational cost is far too expensive when considering potential extreme events for longer return periods and simulation run times on the order of real time. Extrapolation methods, generally based on Weibull distributions, can be explored with a limited dataset. However, this approach requires prior knowledge of the response distribution with particular focus on the tail of the distribution.

Other methods to identify extreme behavior efficiently without overextending assumptions have been developed. One such method is the Design Loads Generator (DLG) (Alford 2008, Kim 2012). DLG was initially developed for linear systems with stochastic Gaussian input, and drew from modified phase distributions based on Extreme Value Theory to generate ensembles of extreme realizations for a given return period.

Another method that has been explored is a lower-fidelity simulation tool that retains major nonlinearities to identify extreme conditions, and then running the identified conditions with a higher-fidelity simulation tool (Reed 2021). In this framework, a surrogate model does not need to be identified but requires a high level of correlation at the peaks between the two simulation tools employed.

Extreme event prediction in the ocean space has also been attempted with machine learning methods. In Guth (2023), extreme statistics of the vertical bending moment were estimated with a wave-episode approach. These wave episodes were generated with the Karhunen-Loeve Theorem and then the responses were estimated by reduced-order models created through Gaussian Process Regression.

Wan et al. (2018) introduced an LSTM-based method to predict extreme events in complex dynamical systems. The methodology details an LSTM architecture that provides a reduced-order model to estimate the non-Galerkin contributions to state dynamics of the model of interest.

In this study, a multi-fidelity approach with neural network correction is investigated. A neural network will be trained to correct extreme low-fidelity hydrodynamic simulation results. The goal is to train the network with more limited data while still retaining the ability to correct the lower-fidelity results to produce quantitatively accurate higher-fidelity response in the most extreme cases. The intent is to recover the extreme statistics and information about the specific wave groups that lead to the extreme event.

This study will specifically focus on the pitch response of the Office of Naval Research Tumblehome (ONRT) flared variant hull form in head seas for Sea State 5 (significant wave height of 4.0 meters and modal period of 15.0 seconds,) and Sea State 6 (significant wave height of 6.0 meters and modal period of 12.0 seconds,) and compare the results of the trained neural network with the higher-fidelity simulation tool.

## 2.  APPROACH

To address the challenge of efficient extreme value prediction, a data-adaptive multi-fidelity approach provides a methodology for evaluating the expected peak ship motion responses for given seaways and ship loading conditions. Such a model could provide actionable information for autonomous seakeeping to improve safety of operation in realistic environments. In Levine et al. (2021), a reduced-order model known as SimpleCode was introduced as a potential candidate for automated seakeeping guidance. In SimpleCode, a volume-based algorithm is applied to model the body-nonlinear hydrostatic and Froude-Krylov forces. By simplifying the local variation of wave pressure (i.e. Smith effect in Bertram, 2011), the surface integral can be converted to a volume integral in the equations for hydrostatic and Froude-Krylov forces by employing Gauss theorem (equations 1 through 13 in Weems and Wundrow, 2013).

As a result, the sectional hydrostatic and Froude-Krylov forces need only evaluate the instantaneous submerged volume and its geometric center. These calculations can be implemented to run quickly with pre-computed Bonjean curves for each station, and a triangle correction reflecting the instantaneous roll relative to the wave as shown in Figure 1. The sectional calculation requires:
- Finding an intersection of flat waterline with sides.
- Interpolation of pre-computed Bonjean curve of two points.
- Calculation of correction values for the area and moments depicted in light blue in Figure 1.

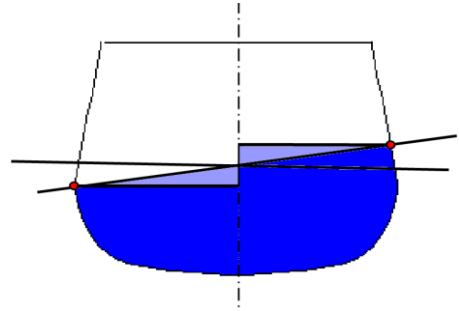

**Figure 1:** Sample sectional volume calculation for the ONR Topsides Series Tumblehome hull.

The complete instantaneous submerged volume and its center is computed by integration of sectional values over the hull shown in Figure 2.

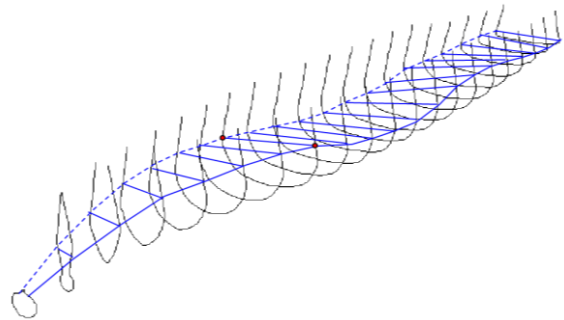

**Figure 2:** Station and incident wave intersection points for ONR Tumblehome hull in stern oblique seas.

The resulting software implementation can support faster than real-time (2,000-5,000x faster) simulations (Weems and Wundrow 2013).

While SimpleCode offers a significant advantage in processing speed, a fidelity gap still exists in the estimation of nonlinear responses as compared to predictions from a higher-fidelity but computationally more intensive simulation tool such as the Large Amplitude Motion Program (LAMP) (Shin et al. 2003; Lin et al. 1990).

In LAMP, a potential flow time-domain approach is used for arbitrary large-amplitude motions of a surface-piercing body in a seaway. The exact body boundary condition is applied on the instantaneous submerged hull surface while a linearized free-surface condition is used. This approximation can be justified in principle upon the assumptions of small incident wave slopes and slenderness of the body geometry in the directions of the (large-amplitude) motions.

In a boundary-element approach, the submerged body surface at each time step is divided into a number of panels over which linearized transient free-surface sources are distributed. The problem is formulated in a coordinate system fixed in space. This is necessary for the case of arbitrary large-amplitude motions and excursions. Under this formulation, a general and concise waterline integral term can be derived to account for arbitrary translations and distortions of the body waterplane, and the diffraction problem can be included in a straightforward manner by adding the incident wave contribution to the body boundary condition. For general nonlinear calculations, the position and orientation of the body is updated (by solving the equations of motion or as prescribed) and the underwater body surface is repanelized at each time step.

In Levine et al, 2024, data-adaptive Long Short-Term Memory (LSTM) neural networks were investigated as part of a multi-fidelity approach incorporating LAMP and SimpleCode. An assessment of this multi-fidelity approach focused on prediction of ship motion responses for roll, pitch and heave in waves. LSTM networks were trained and tested with LAMP motion time-series as the target, and SimpleCode motion time-series and wave time-series as inputs. LSTM networks improve the fidelity of SimpleCode seakeeping predictions relative to LAMP, while retaining the computational efficiency of a lower-fidelity model.

In this paper, the multi-fidelity approach with LSTM is further investigated and applied to the prediction of extreme (transient peak) events. This initial feasibility study specifically focuses on the peaks (maxima) of pitch motion response, which can help inform safe operational guidance.

An LSTM network is trained and tested with LAMP motion prediction time-series as a target, and SimpleCode motion predictions and wave time-series as inputs. LAMP-2 is the version for the simulations in this study. LSTM networks are assessed in terms of improving the fidelity of SimpleCode pitch maxima predictions relative to LAMP, while retaining the computational efficiency of a reduced-order model.

Both LAMP and SimpleCode are configured for three-degree-of-freedom (3-DOF) where pitch, roll and heave are free to move while surge, sway and yaw are fixed while moving forward at a constant speed. For the given hull form geometry, wave conditions, and specified ship speed and heading, LAMP generates pseudo-random irregular waves and predicts motion response, which serves as training and testing datasets. The equivalent input wave fields are then utilized in SimpleCode, which predicts a lower-fidelity response time-series. The 3-DOF motion simulations are performed in head sea conditions with no lateral forces acting on the constrained horizontal degrees of freedom. Consequently, the roll response in head seas is negligible.

## 2.1 Long Short-Term Memory

An LSTM network (Hochreiter and Schmidhuber, 1997) is a type of recurrent neural network, which incorporates both short and long-term effects based on data-adaptive learning for estimation of a system, function, or process. The architecture of an LSTM cell including the inputs, outputs, and short and long-term memory components is in Figure 3.

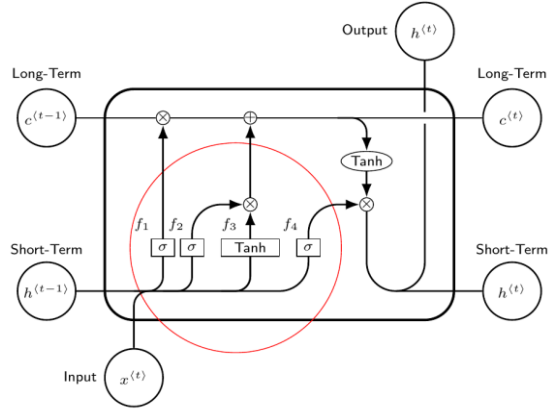

**Figure 3:** Basic architecture of an LSTM cell

Inside the cell, the circled functions labeled ($f_1$, $f_2$, $f_3$, $f_4$) are representative of an LSTM unit, and circled operators are component-based operations. *Sigma* ($\sigma$) is the sigmoid activation function, and $tanh$ is the hyperbolic tangent activation. The size of the weight and bias matrices in the LSTM unit defines the size or dimensionality of the hidden and cell states.

Weights are represented by ($W, U$), biases ($b$), and Hadamard product ($\circ$). Based on the training of the neural network, each of the weights and biases are "learned" through an optimization process. The hidden state ($h_t$) and cell state ($c_t$) act as "memory", and change over time.

The following equations show the compact form of the operations inside of the cell.

$$f_1 = \sigma(W_{f1} \cdot x^{\langle t \rangle} + U_{f1} \cdot h^{\langle t-1 \rangle} + b_{f1}) \quad (1)$$

$$f_2 = \sigma(W_{f2} \cdot x^{\langle t \rangle} + U_{f2} \cdot h^{\langle t-1 \rangle} + b_{f2}) \quad (2)$$

$$f_3 = tanh(W_{f3} \cdot x^{\langle t \rangle} + U_{f3} \cdot h^{\langle t-1 \rangle} + b_{f3}) \quad (3)$$

$$f_4 = \sigma(W_{f4} \cdot x^{\langle t \rangle} + U_{f4} \cdot h^{\langle t-1 \rangle} + b_{f4}) \quad (4)$$

$$c_t = f_1 \circ c^{\langle t-1 \rangle} + f_2 \circ f_3 \quad (5)$$

$$h_t = f_4 \circ tanh(c^{\langle t \rangle}) \quad (6)$$

Function ($f_1$) is the "forget gate", which controls the parts of the long-term state that are deleted. Function ($f_3$) represents the input gate, which controls the information from ($f_2$) that is added to the long-term state. Function ($f_4$) controls the output gate, which determines the output of the long-term state.

By constructing layers of cells, the LSTM network is capable of forecasting a desired response to a provided input. An LSTM network effectively maps the input time-series to the desired output time-series or target(s). Adding more units to each cell or increasing the number of layers can enable the network to model more complex interactions or behaviors, but at greater computational cost.

The training process as well as the accuracy and deficiency of an LSTM network depend on the selection of hyperparameters. The accuracy and time-efficiency of an LSTM network depend on the hyperparameters. The hyperparameters are comprised of the number of inputs, number of network layers, training data sequence length, time resolution of the input time-series, hidden state size, bi-directionality, and dropout method. The length of the training data sequences is equivalent to the number of samples of the input time-series.

Time resolution is based on uniform resampling of the input and target time-series. Through resampling, each time-series is reconfigured into a matrix of size $[N/\tau, \tau]$; where, $N$ is the original time-series length, and $\tau$ is the time resolution factor. If $N$ is not divisible by $\tau$, then the time-series is reduced in length to the closest multiple of $\tau$.

The network includes layers of LSTM cells that contain the hidden states. The hidden state size is the number of parameters ($h_t$), which affects the number of weights in an LSTM unit. Size of the weights vectors are $W \in R^{hxd}$ and $U \in R^{hxd}$ for hidden state size ($h$), and number of input time-series channels ($d$).

The basic architecture of the LSTM framework in this study is in Figure 4.

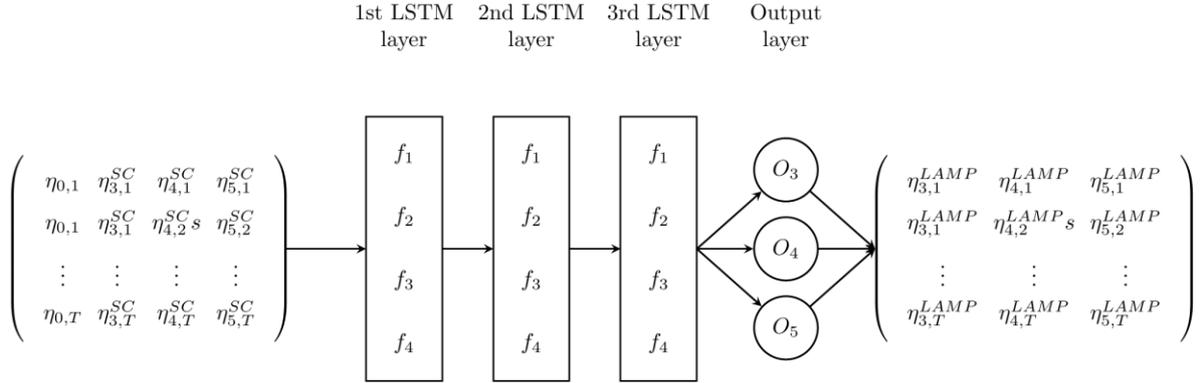

**Figure 4:** LSTM Architecture

An additional fully connected linear output layer has been inserted after the three LSTM layers to parse the output into individual time-series channels for the 3-DOF motions for roll, pitch and heave. Also the labels "SC" and "LAMP" indicate the source of the time-series values. Definitions of the parameters are in Table 1.

**Table 1:** Parameters for LSTM Architecture

| Parameter Definition | Variable |
|---|---|
| Input wave at time-step $j$ | $\eta_{o,j}$ |
| Total number time steps | $T$ |
| $i^{th}$ degree-of-freedom at time-step $j$ | $\eta_{i,j}$ |
| $k^{th}$ gate for LSTM layer | $f_k$ |
| Number of LSTM units per layer | $n$ |
| Output layer cell for DOF $m$ | $O_m$ |

The LSTM framework is comprised of two-stages with SimpleCode as the first stage and LSTM network as the second. In this architecture, four time-series channels are applied as inputs to the LSTM network.

The input channels encompass the 3-DOF motion responses from SimpleCode, and corresponding input wave time-series data. The output response time-series from the two-stage LSTM-SimpleCode model are then compared to 3-DOF motion time-series target from LAMP. The result of training is a set of weights and biases that can be insert directly into Equations 1-6 or into LSTM configurations in existing machine learning packages such as PyTorch in python.

## 2.2 Case Study

In this study, the hull form geometry and loading conditions are based on a model of the ONRT flared hull variant configuration from the ONR Topsides series (Bishop et al. 2005). A rendering of the hull for the model is in Figure 5. The particulars are shown in Table 2.

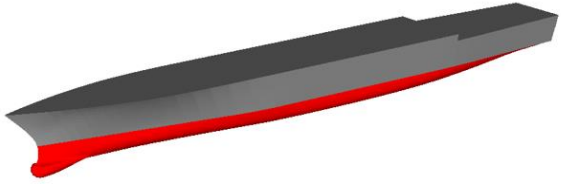

**Figure 5:** Rendering of ONRT flared hull variant.

**Table 2:** ONRT Flared Hull Variant Particulars

| Particular | Value |
|---|---|
| Length ($L_{BP}$) | 154.0 m |
| Beam | 18.8 m |
| Draft | 5.5 m |
| Displacement | 8730.0 t |

To assess the performance of the LSTM method, LAMP and SimpleCode time-series were generated for 10,000 realizations in head seas at a ship speed of 10 knots. Each realization was comprised of 19,000 time samples, including a wave ramp-up length of 1,000 points, and sampled at 10 Hz. The model was free in the vertical plane for heave, roll and pitch, but constrained to constant course and speed in the horizontal plane for surge, sway and yaw. Since the model is operating in head seas, the roll motion is negligible and was not included as an input or output channel.

Irregular unidirectional waves were generated based on the Longuet-Higgins model (Longuet-Higgins, 1984) with a Bretschneider spectrum with significant wave height ($H_s$) of 4.0 m, and modal period ($T_m$) of 15.0 seconds.

The input time-series were rearranged into $[n, N/\tau, \tau]$ arrays by number of inputs $(n)$, number of points per realization $(N)$, and time resolution factor $(\tau)$. Each LSTM layer consisted of a single LSTM cell with its own set of gates ($f_1, f_2, f_3, f_4$), distinct weights, and biases.

The training values for the hyperparameters are in Table 3.

**Table 3:** Training Values for Hyperparameters

| Hyperparameter | Value |
|---|---|
| Time Resolution Factor | 9 |
| Hidden State Size | 30 |
| Number of LSTM Layers | 2 |
| Learning Rate | 0.01 |

For the 10,000 LAMP simulations, 500 were targets for training, 500 for validation, and the remaining 9,000 realizations were for testing of the LSTM network.

The objective function for training was the mean-squared error (MSE) between LAMP and the LSTM output. Equation for MSE is given by:

$$MSE = \frac{1}{N} \sum_{I=1}^{N} \left( y_{LAMP}(t_i) - y_{LSTM}(t_i) \right)^2 \quad (7)$$

In this equation, LAMP time-series is represented by $y_{LAMP}(t)$, response time-series from LSTM network output is represented by $y_{LSTM}(t)$, number of samples in time-series $(N)$, and time index $t_i$.

In Levine et al. (2024), training of LSTM-SimpleCode is applied to long time-series epochs of the 30 minutes per realization. This enabled robust prediction of the signal significant amplitude of the pitch, heave and roll response for a given seaway condition. The general methodology applied in Levine et al. (2024) will be referred to as "Base LSTM" in this study.

The training of the LSTM framework was performed using an NVIDIA Quadro T2000 graphics processing unit (GPU).

In this study, the primary focus is improving the prediction of extreme pitch response both in terms of both the amplitude and time phase estimation of the peak events. The large pitch events identified in LAMP will be considered the extreme targets. It can be noted that the generation of waves in both LAMP and SimpleCode is through uni-directional wave spectra and a transformation into the time domain through linear superposition of the frequency components. In reality, the ocean will likely consist of more directional spreading and steeper (likely breaking) waves which would ultimately result in different responses. That being said, the largest responses in an ocean environment are likely the result of a more focused wave direction. Given the uni-directional focusing of the waves and non-linear considerations of LAMP, the pitch responses generated in LAMP will be considered the "truth" with the understanding that the reality could be different. Still, the presented framework is modular and could potentially be trained on higher-fidelity or experimental time series data.

The method presented seeks to improve the extreme event estimation capability of LSTM networks. Both reduced-order SimpleCode and Base LSTM generally under-predict motion amplitudes relative to LAMP, especially for the largest peaks. The SimpleCode time-series can also exhibit time phase offsets relative to LAMP.

The transient and rare nature of extreme events presents a specific challenge in machine learning in terms of providing sufficient number of realizations of peak behavior for training. To address this challenge, training data is formulated in terms identifying short time windows surrounding the largest peaks in the SimpleCode and LAMP time-series.

Short time windows were identified to extract a "snippet" of time samples around each significant pitch peak. Instead of training on the entirety of a realization, the network was trained with the pitch snippets identified in SimpleCode, along with the heave and wave elevation at the same time steps as the pitch snippets, as input and the corresponding LAMP pitch snippet as the target. In this preliminary study, the snippet time window consisted of 50 seconds centered around each selected SimpleCode peak. Given the relative wave direction and modal period of 15.0 seconds, a time window of 50 seconds allows for, on average, around two wave encounters before and after the large pitch event of interest. The inclusion of these wave encounters not only allows for phase differences between the SimpleCode and LAMP pitch events of interest but also includes valuable information in the behavior of the pitch response leading up to and following a large event. This information can improve the transformation the LSTM framework provides.

An additional test was performed to identify the number of SimpleCode peaks that are considered significant from each realization. While SimpleCode and LAMP share a substantial level of correlation, the largest SimpleCode peak and largest LAMP peak do not always line up in time.

A single LSTM network was generated based on training in Sea State 5 head seas. In addition to Sea State 5, the LSTM network trained with solely Sea State 5 data was tested further to assess robustness for a case in Sea State 6.

For the neural network, the maximum number of training epochs was 1000. The training was considered complete when the maximum number of epochs was achieved, or if the mean-squared error of the training data did not improve by at least 1 percent over a set of 100 epochs.

After training was completed, the test realizations were used for providing an unbiased evaluation of the model fit.

### 2.3 Initial Performance Assessment

The results of the SimpleCode-LAMP correlation test to identify the necessary number of SimpleCode peaks per realization are shown first. Then, using the results to train the network, the time-series maxima generated from the snippet-based LSTM method were compared against LAMP, SimpleCode, and the LSTM method in Levine et al. (2024). Following the comparison, the Sea State 5 trained network was further applied to a case in Sea State 6.

### 3. RESULTS

### 3.1 SimpleCode and LAMP Comparison

SimpleCode and LAMP share a significant level of correlation due to the 3-DOF simulations where the surge and sway of the models are identical along with the experienced wave elevation at the ship's center of gravity. However, the ship responses predicted by lower-fidelity SimpleCode differs from LAMP particularly as the incident waves become more nonlinear. In general, peak amplitude predictions by SimpleCode under-predict likely due to the volume integral approach as opposed to the direct pressure integration in the corresponding LAMP peaks as shown in Figure 6.

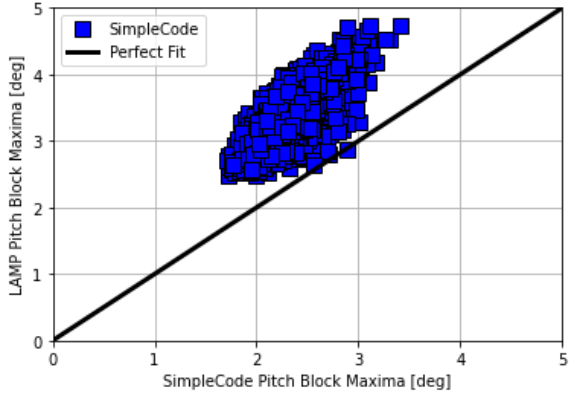

**Figure 6:** Pitch time-series maxima comparison between SimpleCode and LAMP.

Each data point in Figure 6 is indicative of an identical incident wave realization. To measure the correlation, the correlation coefficient was as defined in Equation 8.

$$\rho = \frac{cov(x,y)}{\sigma_x \sigma_y} \quad (8)$$

Correlation coefficient is the ratio between the covariance of two random variables $x$ and $y$, and product of the respective standard deviations $\sigma_x$ and $\sigma_y$. The correlation coefficient between SimpleCode and LAMP is 0.763.

To account for the phase alignment between LAMP and SimpleCode, the peaks were ordered chronologically, and the corresponding LAMP and SimpleCode peaks were compared. The relative rank of the SimpleCode pitch peak that is related to the largest LAMP time-series maxima pitch peak is in Figure 7. The relative rank is defined as the amplitude-ranked value of the SimpleCode peak that corresponds to the largest LAMP peak for a given realization.

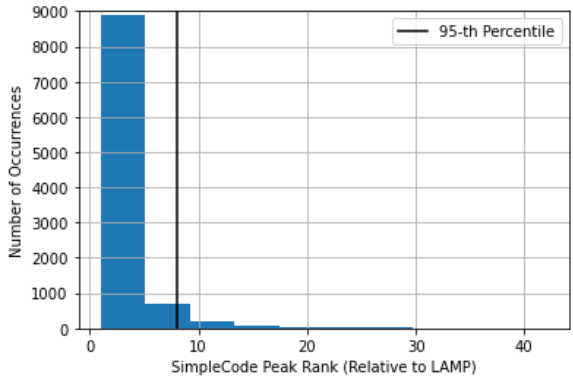

**Figure 7:** Relative rank of SimpleCode pitch peak versus LAMP time-series maxima for each realization.

A value of 1 on the horizontal axis of Figure 7 signifies that the largest SimpleCode peak occurred at the same time instant as the LAMP time-series maxima for that realization while a value of 10 indicates that the 10th largest SimpleCode pitch peak from that given realization occurred at the same time instant as the LAMP time-series maxima. For nearly 90% of realizations, the largest SimpleCode peak and largest LAMP peak occurred at corresponding time instants. While capturing the LAMP time-series maxima is important, including up to 30 snippets per realization in training could begin to include more standard LAMP pitch peaks instead of only the largest values, perhaps over-generalizing the method. So, a threshold of 95% was chosen as a balance between completeness and over-generalization. As a result, the LSTM method used 8 snippets per realization in the training process.

In the testing stage, the 8 snippets were extracted from new SimpleCode time-series. After running each snippet through the trained LSTM, the resulting output with the largest pitch peak was selected for analysis.

### 3.2 LSTM, SimpleCode and LAMP Comparison

To analyse the performance of the LSTM snippet-based approach, time-series maxima were collected from each of the 9,000 test realizations. Then, a probability density function (PDF) was fit by kernel density estimation. The resulting PDFs for SimpleCode, the LSTM method in Levine et al. (2024) ("Base LSTM"), Snippet LSTM and LAMP are in Figure 8.

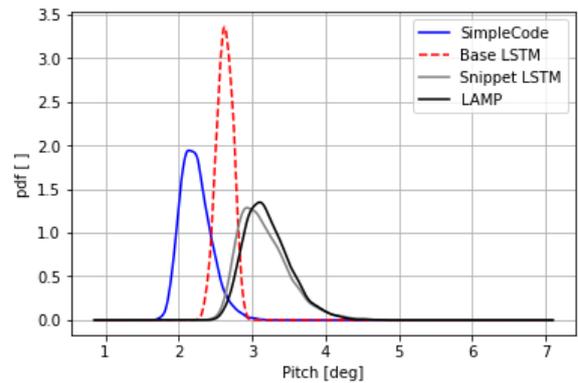

**Figure 8:** Pitch time-series maxima probability density functions for SimpleCode, Base LSTM, Snippet LSTM, and LAMP.

The LSTM methods provide an immediate improvement compared to SimpleCode with respect to the LAMP results. Base LSTM under-predicts LAMP in both most probable maximum (a 16.1% under-prediction) as well as heaviness of tail. By applying the Snippet LSTM, both the most probable maximum and probability density functions (PDF) shape of LAMP is within 5.5% in the peak location and 0.1% in the location of the 95th percentile. To check how time-series maxima compare across individual realizations, the $R^2$ metric was employed to quantify the accuracy of the methods' estimates of the pitch time-series maxima. Equation 9 defines the $R^2$ metric.

$$R^2 = 1.0 - \frac{\sum_{i=1}^{n}(y_i - f_i)^2}{\sum_{i=1}^{n}(y_i - \bar{y})^2} \quad (9)$$

Here, $y_i$ indicates a sample of the true value, or the LAMP time-series maxima for a single realization, $f_i$ is the estimation for that sample, or the low-fidelity estimate for the same realization, and $\bar{y}$ is the mean value of all samples of the true values. A score of 1.0 would indicate a perfect prediction of LAMP time-series maxima while negative scores indicate the method is a worse predictor than the mean value of the LAMP time-series maxima.

In Figure 9, the time-series maxima estimated for SimpleCode, Base LSTM and Snippet LSTM are compared with LAMP.

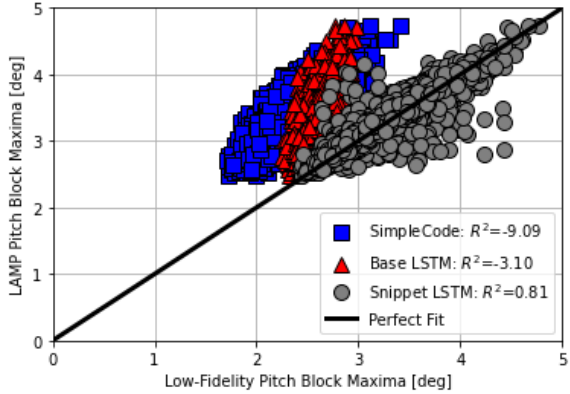

**Figure 9:** Pitch time-series maxima for SimpleCode, Base LSTM, and Snippet LSTM versus LAMP.

The results indicate that Snippet LSTM is an improved predictor of the LAMP time-series maxima compared with SimpleCode and Base LSTM method.

As for capturing specific peak behavior, the time-series can be analyzed individually. The snippet surrounding the largest LAMP pitch event in the ensemble is in Figure 10 along with the corresponding SimpleCode, Base LSTM, and Snippet LSTM pitch time-series.

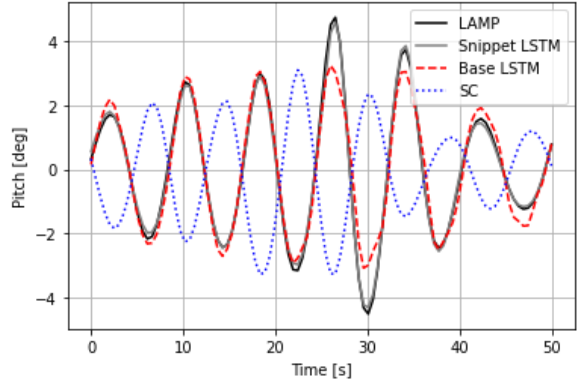

**Figure 10:** Pitch time-series during the largest pitch event predicted by LAMP in the ensemble along with the corresponding SimpleCode, Base LSTM, and Snippet LSTM pitch time-series.

While both the LSTM methods correct the phasing issue evident in the SimpleCode time-series, the snippet-based approach also captures the peak magnitudes (maxima and minima) leading up to and following the peak value. In addition, the peak value is nearly captured (within 4%).

### 3.3 Extension to Sea State 6

To test the flexibility of the Snippet LSTM method, the network trained to correct Sea State 5 SimpleCode simulations was directly applied to Sea State 6 with significant wave height of 6.0 meters and a modal period of 12.0 seconds. The probability density functions of pitch time-series maxima for SimpleCode, the LSTM snippet-based approach with the Sea State 6 SimpleCode simulations as input, and LAMP are in Figure 11.

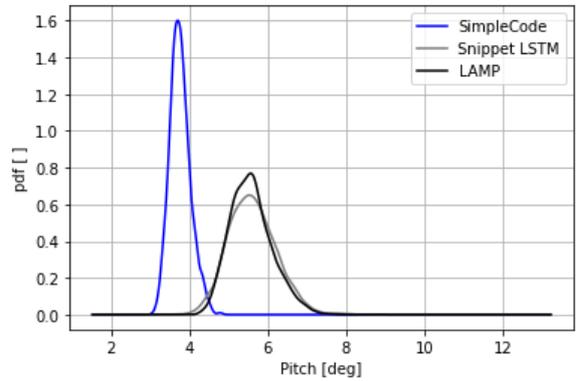

**Figure 11:** Pitch time-series maxima PDFs in Sea State 6 for SimpleCode, Snippet LSTM, and LAMP.

Comparison of Snippet LSTM and LAMP for Sea State 6 resulted in a most probable maximum within 1.5% and the 95th percentile within 0.1%. In addition to the significant wave height increase in Sea State 6, the modal period decreased, resulting in more steep waves and an increased amount of nonlinear dynamics. Without any Sea State 6 data included in the training, the Snippet LSTM appears to identify behavior in SimpleCode time-series and wave elevations while providing corrections comparable to higher-fidelity LAMP for different conditions.

For time-series prediction, the response near the largest pitch event in the LAMP time-series ensemble for the Sea State 6 dataset is in Figure 12.

Even without any Sea State 6 data, the Snippet LSTM approach is able to predict the overall characteristics of the peak event waveform. Ensemble correlation coefficient was 0.71, which is lower as compared to the Sea State 5 result.

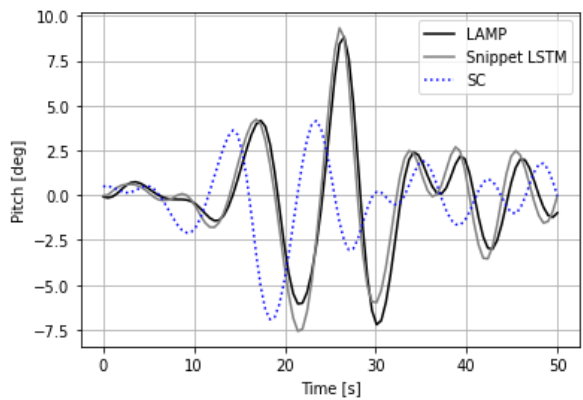

**Figure 12:** Pitch time-series in Sea State 6 for SimpleCode, Snippet-based LSTM, and LAMP.

## 4. CONCLUSIONS

An objective of this study was to assess the potential feasibility of a data-adaptive multi-fidelity model for autonomous seakeeping, particularly the estimation of large pitch events. Data adaptive tuning (or correction) of reduced-order model predictions have been implemented based on training with higher fidelity ship motion response data. From these initial results, this approach may provide a plausible means for improving the performance of a reduced-order model for peak pitch event prediction.

LSTM neural networks have been considered as part of a multi-fidelity approach for prediction of 3-DOF ship motion responses in waves. LSTM networks were trained and tested with LAMP simulations as a target, and SimpleCode simulations and wave time-series as inputs. LSTM networks improve the fidelity of SimpleCode seakeeping predictions relative to LAMP while retaining the computational efficiency of a lower-fidelity simulation tool.

The LSTM neural network trained through a hybrid approach comprised of a physics-based model and data-adaptive stage. The results indicate that Snippet LSTM is an improved predictor of the LAMP time-series maxima compared with SimpleCode and Base LSTM method.

The method appears to be extendable to a higher sea state without any additional training data from the higher sea state. Based on the results of this study, the LSTM trained with the snippet-based approach is a suitable candidate for further investigation and application to extreme event predictions.

Potential future work includes:

- Application of this multi-fidelity approach with LSTM neural networks for automated guidance
- For prediction of 6-DOF motions, structural loads, accelerations, and resistance.
- Extending assessment to cover a range of wave parameters including significant wave heights, modal periods, ship speeds, and relative wave directions.
- Application to other hull form geometries.
- Evaluation of LSTM network configurations in terms of hyperparameters and prediction performance.

## 5. ACKNOWLEDGEMENTS

Authors would like to thank the DoD SMART SEED Grant for sponsoring in-part this study.

Additionally, the authors would like to acknowledge Dr. Vadim Belenky and Kenneth Weems of Naval Surface Warfare Center, Carderock Division; Professor Vladas Pipiras of University of North Carolina; and Professor Themistoklis Sapsis of Massachusetts Institute of Technology for sharing their technical insight.

## 7. DISCUSSION

Armin Troesch
ABS Professor Emeritus of Marine and Design Performance
Deparment of Naval Architecture
University of Michigan

The authors are to be congratulated on presenting a well-written paper investigating the application of a neural network method in the estimation of extreme ship responses. As discussed in the paper, extensive Monte Carlo simulations containing rare extreme events are generally not feasible for complex nonlinear systems. As a result, numerous strategies, such as the one described by the authors, have been proposed and developed where ensembles of short time series containing large responses are constructed.

Using pitch maxima of an ONRT flared hull variant as an example of the method, the authors showed impressive correlation between extreme value PDF's based on Snippet LSTM and LAMP simulations. However, pitch motion is characteristic of a "well behaved" nonlinear process in that pitch nonlinearities become apparent in a "regular" fashion with increasing excitation. This allowed the authors to successfully extend the Sea State 5 results to Sea State 6 without additional training.

1. Would the authors please comment (speculate?) on how they expect the multi-fidelity approach with LSTM would work when applied to nonlinear systems that contain possible bifurcations, e.g. vessel capsize.

The authors also identified potential future applications of the LSTM method.

2. Do they feel that the method is applicable when the nonlinear effect of interest has no SimpleCode equivalent? For example, sonar dome slamming pressures can be estimated using CFD or an impact model in LAMP. How would the authors propose to use the LSTM method on nonlinear processes such as extreme bottom slamming loads?

## 8. AUTHOR'S REPLY

The authors are grateful for Professor Troesch's insights and questions. The responses to the questions are below.

**Question 1:** The application of neural networks to problems identifying extremes necessitates a special approach, as discussed in the paper. The basis of the solution was to make the rare events seem less rare by initially identifying candidates for large pitch events in SimpleCode and then only examining a small "snippet" of time around those candidate rare values. In investigating highly nonlinear systems that may contain bifurcations, the basis for the method would still have to apply. That is to say, there would need to be sufficient data including the different domains of attraction in the training set.

A major crux of this application is whether or not the nonlinear dynamics of such an event would be captured in SimpleCode. SimpleCode does include the most important nonlinear effects in the body-nonlinear hydrostatic and Froude-Krylov forces. The training approach would be relatively straight-forward if SimpleCode was able to qualitatively model these excursions into different dynamical domains.

If SimpleCode was unable to properly model the bifurcations, there are still approaches that would allow for system identification. In Bury et. al (2023), an application identifying the period-doubling bifurcation in chicken heart rates was performed with a deep-learning classifier involving an LSTM. The approach presented involved system inputs that were not directly related as in the SimpleCode-LAMP case, so it stands to reason that with more qualitative knowledge, a similar approach could succeed in the SimpleCode-LAMP framework.

**Question 2:** One major driver in the success of an LSTM network is the relation between the input and output. In the presented case, the high level of correlation between SimpleCode and LAMP resulted in a network that performed well. However, an LSTM network can still perform well even without a process as correlated to LAMP as SimpleCode. In Levine et.al (2024), ship motion statistics generated in LAMP were accurately captured by an LSTM approach that only used the incident wave profile as input. In this data-driven approach, the standard deviations of heave, roll, and pitch were well estimated – most predictions were at least 95% accurate - for various ship speeds, sea states, and relative wave headings.

That being said, it is unlikely this same approach could be directly applied to the rare event prediction application. In this approach, there must be a way to identify when the large pitch events occur in LAMP. To apply this method to nonlinear processes like extreme bottom slamming loads, an input process that is related and has at least some level of correlation in extremes would need to be identified. One possible candidate for the extreme bottom slamming load response could be relative velocity as predicted by SimpleCode. Of course, further study would need to be performed to investigate the level of correlation and number of events that would have to be considered, as in Figure 7.